\documentclass[runningheads]{llncs}
\usepackage{graphicx}
\usepackage{amsmath,amssymb} % define this before the line numbering.
\usepackage{color}
\usepackage{dsfont}
\usepackage{multirow}
\usepackage{slashbox}
\usepackage{tabularx}
\newcolumntype{Y}{>{\centering\arraybackslash}X}
\usepackage{subfig}
\usepackage{url}
\usepackage{enumitem}
\usepackage{array}
\usepackage[width=122mm,left=12mm,paperwidth=146mm,height=193mm,top=12mm,paperheight=217mm]{geometry} 
\newcommand{\specialcell}[2][c]{%
  \begin{tabular}[#1]{@{}l@{}}#2\end{tabular}}
\begin{document}
% \linenumbers
\pagestyle{headings}
\mainmatter
\def\ECCV16SubNumber{34}

\title{Deep Learning the City: \\Quantifying Urban Perception At A
  Global Scale} % Replace with your title

\titlerunning{Deep Learning the City}

\authorrunning{A. Dubey, N. Naik, D.Parikh, R. Raskar, and  C. Hidalgo}

\author{Abhimanyu Dubey\textsuperscript 1, Nikhil Naik\textsuperscript
  3, Devi Parikh\textsuperscript 2 \\ Ramesh Raskar\textsuperscript 3,
 C\'esar A. Hidalgo\textsuperscript 3}

%Please write out author names in full in the paper, i.e. full given and family names. 
%If any authors have names that can be parsed into FirstName LastName in multiple ways, please include the correct parsing, in a comment to the volume editors:
%\index{Lastnames, Firstnames}
%(Do not uncomment it, because you may introduce extra index items if you do that...)

\institute{\textsuperscript 1 Indian Institute of Technology Delhi\\
	\email{abhimanyu1401@gmail.com}\\
	\textsuperscript 2 Virginia Tech\\
	\email{ parikh@vt.edu}\\
	\textsuperscript 3 MIT Media Lab\\
	\email{ \{naik,raskar,hidalgo\}@mit.edu}\\
}

\maketitle

\begin{abstract}

% Describe problem, how we solved it, technical contribution, 
% do we need to compare with hand-crafed feauture-based learner?
Computer vision methods that quantify the perception of urban environment are 
increasingly being used to study the relationship between a city's
physical appearance and the behavior and health of its residents. Yet,
the throughput of current methods is too limited to quantify the
perception of cities across the world. To tackle this challenge, we
introduce a new crowdsourced dataset containing 110,988 images from 
56 cities, and 1,170,000 pairwise comparisons provided by 81,630
online volunteers along six perceptual attributes: safe, lively, 
boring, wealthy, depressing, and beautiful. Using this data, we train
a Siamese-like convolutional neural architecture, which learns from 
a joint classification and ranking loss, to predict human judgments of 
pairwise image comparisons. Our results show that crowdsourcing 
combined with neural networks can produce urban perception data 
at the global scale. 

\keywords{Perception, Attributes, Street View, Crowdsourcing}
\end{abstract}

\section{Introduction}
% intro about problem's importance
% figure that has dataset examples 
% summary of the problem
% summarize contributions
\textit{We shape our buildings, and thereafter our buildings shape
  us.} -- Winston Churchill.\vspace{0.3cm}\\
These famous remarks reflect the widely-held belief among
policymakers, urban planners and social scientists that the physical
appearance of cities, and it's perception, impacts the behavior and health of their
residents. Based on this idea, major policy 
initiatives---such as the New York City
``Quality of Life Program''---have been launched across the world to
improve the appearance of cities. Social scientists have either
predicted or found evidence for the impact of the perceived unsafety
and disorderliness of cities on
criminal behavior~\cite{wilson1982broken,keizer2008spreading},
education~\cite{milam2010perceived},
health~\cite{cohen2003neighborhood}, and
mobility~\cite{piro2006physical}, among others. However, these studies
have been limited to a few neighborhoods, or a handful of cities at
most, due to a lack of quantified data on the perception of
cities. Historically, social scientists have collected this data 
using field surveys~\cite{sampson2012great}. In the past 
decade, a
new source of data on urban appearance has emerged, in the form of
``Street View'' imagery. Street View has enabled researchers to
conduct virtual audits of urban
appearance, with the help of
trained experts~\cite{miller2013using,hwang2014divergent} or 
crowdsourcing~\cite{salesses2013collaborative,quercia2014aesthetic}.   

However, field surveys, virtual audits and crowdsourced studies 
lack both the resolution and
the scale to fully utilize the global corpus of Street View imagery. 
For instance, New York City alone has roughly one million street
blocks, which makes generating an exhaustive city-wide dataset of urban appearance a daunting task. 
Naturally, generating urban appearance data through
human efforts for hundreds of
cities across the world, at several time points, and across different
attributes (e.g., cleanliness, safety, beauty), 
remains impractical. The
solution to this problem is to develop computer vision 
algorithms---trained with human-labeled data---that
conduct automated surveys of the built environment at street-level
resolution and global scale.

A notable example of this approach is Streetscore by Naik et
al.~\cite{naik2014streetscore}---a computer vision algorithm 
trained using Place Pulse 1.0~\cite{salesses2013collaborative}, a
crowdsourced game. In Place Pulse 1.0, users are asked to select one of
the two Street View images in response to question ``Which place looks
safer?'', ``Which place looks more unique?'', and ``Which places looks
more upper class?''. This survey collected a total of  
200,000 pairwise comparisons across the three attributes for 4,109 
images from New York, Boston, Linz, and Saltzburg. 
Naik et al. converted the pairwise comparisons for perceived safety to
ranked scores and trained a regression algorithm using generic image
features to predict the ranked score for perceived safety 
(also see the work by Ordonez and Berg~\cite{ordonez2014learning} and
Porzi et al.~\cite{porzi2015predicting}).  
Streetscore was employed to automatically generate a dataset of 
urban appearance covering 21 U.S. cities~\cite{naik2016cities}, 
which has been used to identify the impact of historic preservation
districts on urban appearance~\cite{been2015preserving}, for
quantifying urban change using time-series street-level 
imagery~\cite{NBERw21620}, and to determine the effects of urban
design on perceived safety~\cite{harvey2015effects}.
%in multiple social science
%studies~\cite{been2015preserving,NBERw21620,harvey2015effects}.

Yet the Streetscore algorithm is not unboundedly scalable. 
Streetscore was trained using a dataset containing a few thousand 
images from New York and Boston, so it cannot accurately 
measure the perceived safety of images from cities outside of the 
Northeast and Midwest of United States, which may have 
different architecture styles and urban planning constructs. This 
limits our ability to generate a truly \textbf{global dataset of 
urban appearance}. Streetscore was also trained  
using a dataset with a relatively dense set of preferences (each 
image was involved in roughly 30 pairwise comparisons). But 
collecting such a dense set of preferences with crowdsourcing is 
challenging for a study that involves hundreds of thousands of 
images from several cities, and multiple attributes. So scaling up 
the computational methods to map urban 
appearance from the regional scale, to the global scale, requires 
methods that can be trained on larger and sparser datasets---which 
contain a large, visually diverse set of images with relatively 
few comparisons among them.

With the motivation of developing a global dataset of urban 
appearance, in this paper, we introduce a new crowdsourced dataset 
of urban appearance and a computer vision technique to rank 
street-level images for urban appearance in this paper. 
Our dataset, which we call 
the \textbf{Place Pulse 2.0} dataset, contains 1.17 million pairwise
comparisons for 110,988 images from 56 cities from 28 countries across
6 continents, scored by 81,630 online volunteers, along six 
perceptual dimensions: safe, lively, boring, wealthy, depressing, and
beautiful. We use the Place Pulse 2.0 (PP 2.0) 
dataset to train convolutional
neural network models which are able to {predict} 
the pairwise comparisons for perceptual attributes by 
taking an image pair as input. We propose two related network 
architectures: (i) the Streetscore-CNN (\textbf{SS-CNN} for
short) and (ii) the Ranking SS-CNN (\textbf{RSS-CNN}). The  
SS-CNN consists of two disjoint identical sets of layers
with tied weights, followed by a fusion sub-network, which minimizes
the classification loss on pairwise comparison prediction. The
RSS-CNN includes an additional ranking sub-network, 
which tries to simultaneously minimize the loss on both
pairwise classification and ordinal ranking over the dataset. 
The SS-CNN architecture---fine-tuned with the 
PP 2.0 dataset---significantly outperforms the same network
architecture with pre-trained AlexNet~\cite{krizhevsky2012imagenet},
PlacesNet~\cite{zhou2014learning}, or VGGNet~\cite{simonyan2014very}
weights. RSS-CNN shows better prediction performance than 
SS-CNN, owing to end-to-end learning 
based on both classification 
and ranking loss. Moreover, our CNN architecture obtains much better
performance over a geographically disparate test set 
when trained with PP 2.0, in
comparison to PP 1.0, due to the larger size and
visual diversity (110,988 images from 56 cities, versus 4,109 images
from 4 cities).  

We find that networks trained to predict one visual 
attribute (e.g., \textit{Safe}), are fairly accurate in the prediction
of other visual attributes (e.g., \textit{Lively}, \textit{Beautiful},
etc). We also use a trained network to predict the perceived safety 
of streetscapes from 6 new cities from 6 continents, 
that were not part of the training set. Finally, we hope that this work
will enable further progress 
on global studies of the social and economic effects of architectural
and urban planning choices. 

\section{Related Work}

Our paper speaks to four different strands of the academic literature: 
(1) predicting perceptual responses to images, (2) using urban imagery 
to understand cities, (3) understanding the connection between urban
appearance and socioeconomic outcomes, and (4) generating image rankings and comparisons.

There is a growing
body of literature on \textbf{predicting the perceptual responses to
images}, such as aesthetics~\cite{joshi2011aesthetics},
memorability~\cite{isola2011makes},
interestingness~\cite{dhar2011high}, and
virality~\cite{deza2015understanding}. In particular, our work is
related to the literature on predicting the perception of street-level
imagery. 
Naik et al.~\cite{naik2014streetscore} use generic image 
features and support vector regression to develop Streetscore, an
algorithm that predicts the perceived safety
of street-level images from United States, using training data
from the Place Pulse 1.0
dataset~\cite{salesses2013collaborative}.  
Ordonez and Berg~\cite{ordonez2014learning} use the Place Pulse 1.0
dataset and report similar results for prediction of perceived safety,
wealth, and uniqueness using Fisher vectors and DeCAF
features~\cite{donahue2013decaf}. Porzi et
al.~\cite{porzi2015predicting} identify the mid-level visual
elements~\cite{doersch2012makes} that contribute to the perception of
safety in the Place Pulse 1.0 dataset. 

This new body of literature that utilizes \textbf{urban imagery to 
understand cities} has been enabled by new sources of data  
from both commercial providers (e.g., Google Street View) 
and photo-sharing websites (e.g., Flickr). These data
sources have enabled applications for computer
vision techniques in the fields of architecture, urban planning, urban
economics and sociology. Doersch et al.~\cite{doersch2012makes}
identify geographically distinctive visual elements from Street View
data. Lee et al.~\cite{linking2015iccp} extend this work in the
temporal domain by identifying architectural elements which are
distinctive to specific historic periods. Arietta et
al.~\cite{arietta2014city} and Glaeser et al.~\cite{glaeser2015big} 
develop regression models based on Street
View imagery to predict socioeconomic
indicators. Zhou et al.~\cite{zhou2014recognizing} develop a unique
city identity based on a high-level set of attributes derived from
Flickr images. Khosla et al.~\cite{khosla2014looking} use Street View
data and crowdsourcing to demonstrate that both humans and computers
can navigate an unknown urban environment to locate businesses.   

Our research also speaks to the more traditional stream of 
literature studying the connection between \textbf{urban appearance and 
socioeconomic outcomes} of urban residents, especially health and
criminal behavioral. Researchers have studied the connection between 
the perception of unsafety and alcoholism~\cite{kuipers2012association}, 
obesity~\cite{dulin2013associations}, and the spread of 
STDs~\cite{cohen2003neighborhood}. The 
influential ``Broken Windows Theory (BWT)''~\cite{wilson1982broken} 
hypothesizes that criminal activity is more likely to occur in places 
that appear disorderly and visually unsafe. 
There has been a vigorous debate among scholars on BWT, who have found evidence in 
support~\cite{kelling1997fixing,keizer2008spreading} and against the 
theory~\cite{sampson2001disorder,harcourt1998reflecting}. 
Once again, this is another area where methods to quantify urban 
appearance may illuminate important questions. 

Finally, our work is related to
literature on \textbf{ranking and comparing images} 
based on both semantic and subjective
attributes, or generating metrics for image comparisons. The
concept of ``relative attributes''~\cite{parikh2011relative}---ranking
object/scene types according to different attributes---has been shown
to be useful for applications such as image
classification~\cite{parkash2012attributes} and guided image
search~\cite{kovashka2012whittlesearch}. Kiapour et al.~\cite{kiapour2014hipster} rank images based
on clothing styles using annotations collected from an online game, and
generic image features.  Zhu et al.~\cite{zhu2014mirror} rank
facial images for attractiveness, for generating better portrait
images. Wang et al.~\cite{wang2014learning} introduce a deep ranking
method for image similarity metric computation. Zagoruyko and 
Komodakis~\cite{zagoruyko2015learning} develop a Siamese architecture 
for computing image patch similarity for applications 
like wide-baseline
stereo. Work on image perception summarized earlier~\cite{naik2014streetscore,ordonez2014learning,porzi2015predicting}
also ranks street-level images based on perceptual
metrics. 

In this paper, we contribute to these literatures by introducing 
a CNN-based technique to predict human judgments on 
urban appearance, using a global crowdsourced dataset. 
\section{The Place Pulse 2.0  Dataset}
\label{sec:Evaluation}
\begin{figure*}[t]
    \centering
    \includegraphics[width=1.0\textwidth]{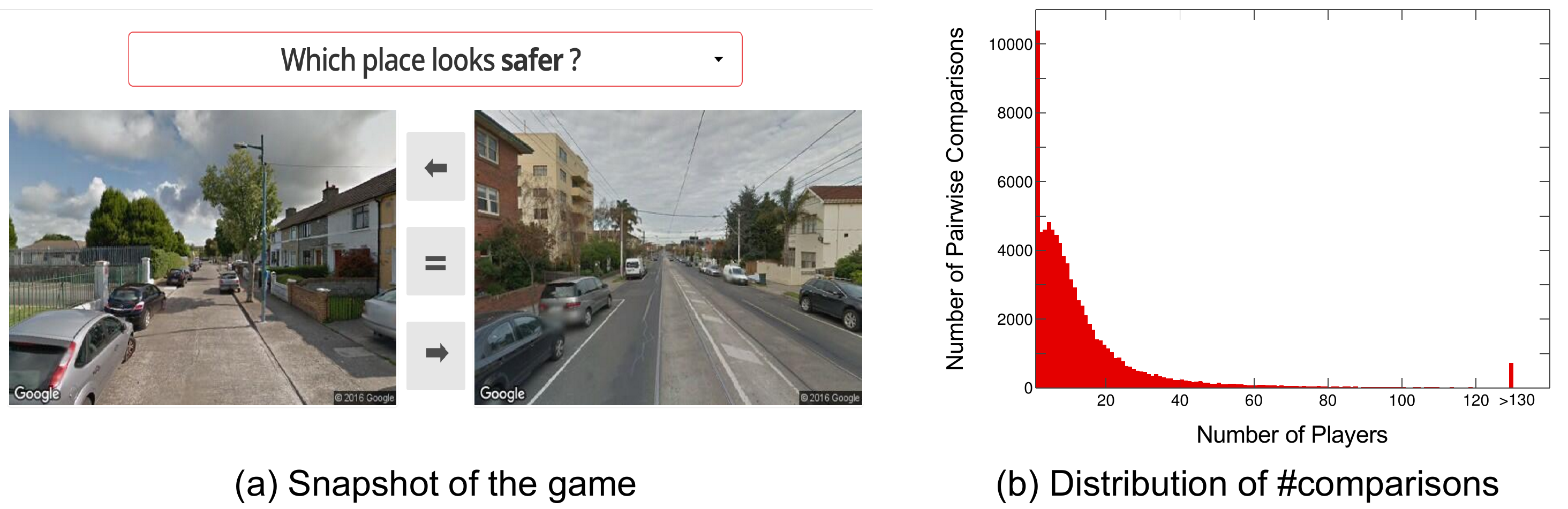} 
   \caption{Using a crowdsourced online game (a), we collect 1.1
     million pairwise comparisons on urban appearance from from 81,630
     volunteers. The distribution of number of pairwise comparisons
     contributed by players is shown in (b).}
\vspace{-0.2cm}
\label{fig:data_summary}
\end{figure*}

Our first goal is to collect a crowdsourced dataset of 
perceptual attributes for street-level images. 
To create this dataset, we chose Google
Street View images from 56 major cities from 28 countries spread
across all six inhabited continents. We obtained the 
latitude-longitude values for locations in these cities 
using a uniform grid~\cite{persson2004simple} of points calculated on 
top of polygons of city
boundaries. We queried the Google Street View Image API\footnote{https://developers.google.com/maps/documentation/streetview/} 
using the latitude-longitude values, and 
obtained a total of 110,988 images captured between years 2007 
and 2012.

Following Salesses et al.~\cite{salesses2013collaborative}, 
we created a web-interface (Figure~\ref{fig:data_summary}-(a)) for
collecting pairwise comparisons from users. Studies have shown that
gathering relative comparisons is a more efficient and accurate way
of obtaining human rankings as compared to obtaining numerical 
scores from each user~\cite{stewart2005absolute,bijmolt1995effects}. 
In our implementation, we showed users a randomly-chosen pair of images
side by side, and asked them to
choose one in response to one of the six questions, preselected by
the user. The questions were: ``Which place looks safer?'', 
``Which place looks livelier?'', ``Which place looks more boring?'', 
``Which place looks wealthier?'', 
``Which place looks more depressing?'', and ``Which place looks 
more beautiful?''.  

We generated traffic on our website primarily from organic media
sources and by using Facebook advertisements targeted to  
English-speaking users who
are interested in online games, architecture, cities, sociology,  and
urban planning. We collected a total of 1,169,078 
pairwise comparisons from 81,630 online users between May 2013 and
February 2016.   
The online users provided 16.6 comparisons on
average. 6,118 users provided a single comparison each, while 30 users
provided more than 1,000 comparisons
(Figure~\ref{fig:data_summary}-(b)). The maximum number of comparisons
provided by a single user was 7,168. We obtained the highest 
responses (370,134) for the question ``Which place looks safer?'',
and the lowest responses (111,184) for the question 
``Which place looks more boring?''.  
We attracted users from 162 countries (based on data from web 
analytics). Our user base contained a good mix of
residents of both developed and developing countries. The top five 
countries of origin for these users were United States
($31.4\%$), India ($22.4\%$), United Kingdom ($5.8\%$), Brazil
($4.6\%$), and Canada ($3.6\%$). It is worth noting that the Place
Pulse 1.0 study found that individual preferences for urban appearance
were not driven by participants' age, gender, or
location~\cite{salesses2013collaborative}, indicating that there is no
significant cultural bias in the dataset. Place Pulse 1.0 also found
high inter-user reproducibility and high transitivity in people's
perception of urban appearance, which is indicative of consistency in
data collected for this task. With that established, we did not
collect demographic information from users for our much larger PP 2.0
dataset, but we did use the exact same data collection interface and
user recruitment strategy as PP 1.0. 
Table~\ref{tab:data_summary} summarizes the key facts about the Place Pulse 2.0 dataset. 
% training data
% evaluation metrics
% comparison of features
% visualize filters of convolutional layers
% add ranking examples

\begin{table}[t]
  \centering
  \caption{The Place Pulse 2.0  Dataset at a Glance}
  \subfloat[Statistics on Images]{%
    \hspace{.5cm}%
        \begin{tabular}{lcc}
          \hline\noalign{\smallskip}
          Continent & \#Cities & {\#Images}\\
          \noalign{\smallskip}
          \hline
          \noalign{\smallskip}
          {Asia} & 7 & 11,342 \\
          {Africa} & 3 & 5,069 \\
          {Australia} & 2 & 6,082 \\
          {Europe} & 22 & 38,636 \\
          {North America} & 15 & 33,691 \\
          {South America} & 7 & 16,168  \\ 
          \hline 
          {Total} & 56 & 110,988  \\
          \hline
        \end{tabular}%
    \hspace{.5cm}%
  }
  \subfloat[Statistics on Pairwise Comparisons (PC)]{%
    \hspace{.5cm}%
        \begin{tabular}{lcc}
          \hline\noalign{\smallskip}
          Question & \#PC  & 
          \specialcell{\#Per-image PC}\\
          \noalign{\smallskip}
          \hline
          \noalign{\smallskip}
          {Safe} & 370,134 & 7.67 \\
          {Lively} & 268,494 & 5.52  \\
          {Beautiful} & 166,823 & 3.46 \\
          {Wealthy} & 137,688 & 2.87 \\
          {Depressing}\hspace{0.2cm} & 114,755 & 2.47 \\
          {Boring} & 111,184 & 2.40\\
          \hline
          {Total} & 1,169,078 & 16.73  \\
          \hline
        \end{tabular}%
    \hspace{.5cm}%
  }
\label{tab:data_summary}
\end{table}
      
\section{Learning from the Place Pulse 2.0 Dataset}
We now describe how we use the Place Pulse 2.0 dataset to
train a neural network model to {predict} pairwise comparisons. 
Collecting pairwise comparisons has been the method of choice for
learning subjective visual attributes such as style, perception, 
and taste. Examples include learning clothing 
styles~\cite{kiapour2014hipster},
urban appearance~\cite{naik2014streetscore}, 
emotive responses to GIFs~\cite{jou2014predicting}, or affective 
responses to paintings~\cite{sartori2015affective}. 
All these efforts use a two-step process for learning these 
subjective visual attributes---image ranking, followed by image
classification/regression based on the visual attribute. 
In the first step, these methods~\cite{naik2014streetscore,kiapour2014hipster,jou2014predicting,sartori2015affective} convert the pairwise comparisons to ranked scores for images using the Microsoft 
TrueSkill~\cite{herbrich2006trueskill} algorithm. TrueSkill is a
Bayesian ranking method, which generates a ranked score for a player
(in this case, an image) in a two-player game by iteratively updating
the ranked score of players after every contest (in this case, a
human-contributed pairwise comparison). Note that this approach for
producing image rankings does not take image features into account. 
In the next step, the ranked scores, along with image features are
used to train classification or regression algorithms, to predict the
score of a previously unseen image. 

However, this two-step process has a few limitations. 
First, for larger datasets, the number of 
crowdsourced pairwise comparisons required becomes quite large. 
TrueSkill needs 24 to 36 comparisons per image for 
obtaining stable rankings~\cite{herbrich2006trueskill}. Therefore, 
we would require \textbf{$\sim$1.2 
to 1.9 million comparisons} per question, to obtain stable TrueSkill
scores for 110,988 images in the Place Pulse 2.0 dataset. 
This number is hard to achieve, even with
the impressive number of users attracted by the Place Pulse
game. Indeed, we are able to collect only 3.35 comparisons per
image per question on average, after 33 months of data collection. 
Second, this two-step process ignores the visual content of images
in the ranking process. We believe it is better to use visual content 
in the image ranking stage itself by learning to predict 
pairwise comparisons directly, which is similar in spirit to learning 
ranking functions for semantic attributes 
from image data~\cite{parikh2011relative} (also see 
Porzi et al.~\cite{porzi2015predicting} for additional discussion 
on ranking versus regression). To address both problems, we propose to 
predict pairwise comparisons by training a neural network directly 
from image pairs and their crowdsourced comparisons from the Place Pulse 2.0 dataset. We describe the problem formulation and our neural network model next.\\\\
\textbf{Problem Formulation:} The Place Pulse 2.0 
dataset consists of a set of $m$ images $I =
\{\mathrm{x}_i\}^m_{i=1} \in \mathbb{R}^n$ in pixel-space and a set
of $N$ image comparison triplets $P = \{ (i_k,j_k,y_k) \}^N_{k=1}, \
i,j \in \{1,...,m\}, y \in \{+1,-1\}$, which specify a pairwise
comparison between the $i$th and the $j$th image in the set. 
$y = +1$ denotes
a win for image $i$, and $y=-1$ denotes a win for image $j$. Our goal
is to learn a ranking function $f_r(\mathrm{x})$ on the raw
image pixels such that we satisfy the maximum number of constraints
\begin{eqnarray}
y \cdot (f_r(\mathrm{x}_{i}) - f_r(\mathrm{x}_{j})) > 0 \ \ \forall \
(i,j,y) \in P
\label{eq:1}
\end{eqnarray}
over the dataset. We aim to approximate a solution for 
this NP-hard problem~\cite{joachims2002optimizing} 
using a ranking approach, motivated by the
direct adaptation of the RankSVM~\cite{joachims2002optimizing}
formulation by Parikh and Grauman~\cite{parikh2011relative}. 

As the first step towards solving this problem, we transform 
the ranking task to a 
classification task. Specifically, our goal is to design a function 
which given an image pair, extracts low-level and mid-level 
features for each image as well as
higher-level features discriminating the pair of images, and then
predicts a winner. We next describe a convolutional neural network
architecture which learns such a function.
%Therefore, we learn 
%the function $h_l(\mathrm{x}_i,\mathrm{x}_j)$, which is the 
%probability estimate of the left image winning---with 
%information on both networks present to each function. 
%We output the predicted class $\hat y$ as
%\begin{eqnarray}
%\hat y = {\arg \max}_{x \in \{1,-1\}}(h_x(\mathrm{x}_i,\mathrm{x}_j)).
%\end{eqnarray}\\

\subsection{Streetscore-CNN}
\begin{figure*}[t]
    \centering
    \includegraphics[width=0.9\textwidth]{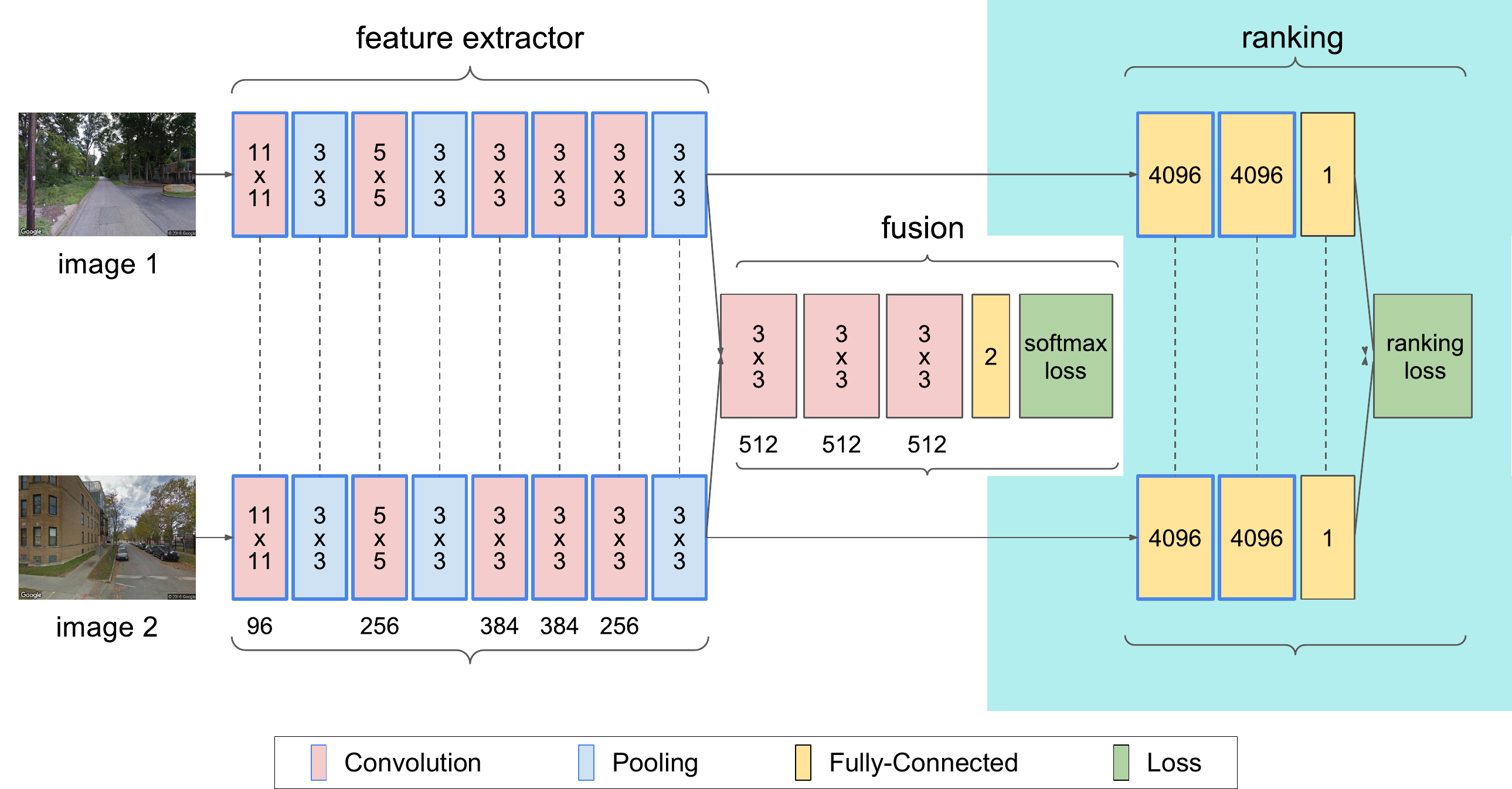}
   \caption{We introduce two networks architectures, based on
     the Siamese model, for predicting pairwise comparisons of urban
     appearance. The basic model (SS-CNN) is trained with softmax
   loss in the fusion layer. We also introduce ranking loss layers  
   to train the Ranking SS-CNN (additional layers shown in light blue
   background). While we experiment with AlexNet,
   PlacesNet, and VGGNet, this figure shows the AlexNet configuration.}
\label{fig:model}
\end{figure*}

We design the Streetscore-CNN (\textbf{SS-CNN}) for predicting
the winner in a pairwise comparison task, by taking an image pair as
input (Figure~\ref{fig:model}). SS-CNN consists of two disjoint identical sets of layers with tied weights for 
feature extraction (similar to a Siamese network~\cite{chopra2005learning}). 
These feature extractor layers are concatenated and followed by 
a \textit{fusion} sub-network, which consists of a set of convolutional layers 
culminating in a fully-connected layer with softmax loss used to train 
the network. The fusion sub-network was inspired by the \textit{temporal fusion} 
architecture~\cite{karpathy2014} used to learn temporal features 
from video frames. The temporal fusion architecture learns
convolutional filters by combining information from   
different activations in time. We employ a similar tactic to 
learn discriminative filters from pairwise image comparisons.
We train SS-CNN for binary classification using the standard 
softmax or classification loss ($L_c$) with stochastic gradient descent. 
Since we perform classification between two categories (left image, 
right image), the softmax loss is specified as
\begin{equation}
\begin{gathered}
L_c = \sum_{(i,j,y) \in P} \sum^K_k - \mathds{1}[y=k] \log (g_k(\mathrm{x}_i,\mathrm{x}_j)) \\
%g(\mathrm{x}_i,\mathrm{x}_j) = \mathrm{softmax}(h_l(\mathrm{x}_i,\mathrm{x}_j),h_r(\mathrm{x}_i,\mathrm{x}_j)) 
\end{gathered}
\end{equation}
where $K=2$ and $g$ is the softmax of final layer activations.

\if 0
The classification loss is given by 
\begin{equation}
\begin{gathered}
L_c =  \sum_{(i,j,y) \in P} - \hat y \cdot \log 
(g(\mathrm{x}_i,\mathrm{x}_j)) \\
\hat y = \{1,0\} \text{ if } y=+1,\hspace{0.3cm} \hat y = \{0,1\} \text{ if } y=-1 \\
g(\mathrm{x}_i,\mathrm{x}_j) = 
\mathrm{softmax}(h_l(\mathrm{x}_i,\mathrm{x}_j),h_r(\mathrm{x}_i,\mathrm{x}_j)), 
\end{gathered}
\end{equation}
where $(h_l(\mathrm{x}_i,\mathrm{x}_j)/(h_(\mathrm{x}_i,\mathrm{x}_j)$
\fi

\subsection{Ranking Streetscore-CNN}
While the SS-CNN architecture learns to predict pairwise comparisons
from two images, training with logistic loss 
does not account for the ordinal ranking over all
the images in the dataset. Moreover, training for only binary
classification may not be sufficient to train such 
complex networks to understand the fine-grained differences 
between image pairs~\cite{wang2014learning}. 
Therefore, we explicitly incorporate the 
ranking function $f_r(\mathrm{x})$ (Eq.~\ref{eq:1}) 
in the end-to-end learning process, we modify 
this basic SS-CNN architecture by attaching a ranking sub-network,
consisting of fully-connected 
weight-tied layers (Figure~\ref{fig:model}, in light blue).
We call this network the Ranking 
SS-CNN (\textbf{RSS-CNN}). The RSS-CNN learns an additional set of 
weights---in comparison to SS-CNN---for minimizing a ranking loss,
\begin{eqnarray}
L_r  = \sum_{(i,j,y) \in P} \big(\max (0, y \cdot (f_r(\mathrm{x}_{j}) - f_r(\mathrm{x}_{i})) \big)^2.  
\end{eqnarray}
The ranking loss ($L_r$) is designed to penalize the network to satisfy
the constraints of our ranking problem---which is identical to the the
loss function of the 
RankSVM~\cite{joachims2002optimizing,chapelle2010efficient} 
formulation. 
%RSS-CNN learns an explicit ranking function over all images, in
%addition to a discriminative set of fusion filters for the
%classification task, similar to {SS-CNN}. 
To train RSS-CNN, we minimize the loss 
function ($L$), which is a weighted combination of the classification 
(or softmax) loss ($L_c$), and the ranking loss ($L_r$), in the form
$L = L_c(P) + \lambda L_r(P)$.
We set the hyper-parameter $\lambda$ using a grid-search to maximize 
the classification accuracy on the validation set. 
%Thus, {RGP-CNN} simultaneously
%learns individual low and mid-level image features, discriminative
%high-level features, and an explicit ranking function over all
%images.\NNnote{we need to justify this I guess}. 

\section{Experiments \& Results}
After defining SS-CNN and RSS-CNN, we evaluate their performance 
in Section 5.1 and Section 5.2,  
using the $370,134$ pairwise comparisons collected for the 
question ``Which place looks safer?'', since this question 
has the highest number of responses. Results for other 
attributes are described in Section 5.3. 

\subsubsection{Implementation Details} For all experiments, 
we split the set of triplets ($P$) for a given
question randomly in the ratio 65--5--30 for training, 
validation and
testing.  We conducted experiments using the latest stable implementation of the Caffe
library~\cite{caffepaper}. For both SS-CNN and RSS-CNN, we initialized  
the feature extractor layers using the pre-trained model weights of 
the following networks using their publicly available 
Caffe models\footnote{https://github.com/BVLC/caffe/wiki/Model-Zoo} (one at a time): 
(i) the \textbf{AlexNet} image classification 
model~\cite{krizhevsky2012imagenet}, (ii) the
\textbf{VGGNet}~\cite{simonyan2014very} 19-layer image classification
model, and (iii) the \textbf{PlacesNet}~\cite{zhou2014learning} scene
classification model. The weights for 
layers in fusion and ranking sub-networks were
initialized from a zero-mean Gaussian distribution with 
standard deviation 0.01, following~\cite{krizhevsky2012imagenet}. 
 
We trained the models on a single
NVIDIA GeForce Titan X GPU. The momentum was set to 0.9. 
The initial learning rate was set to 0.001. When the validation 
error stopped improving with current learning
rate, we reduced it by a factor of 10, repeating this process 
a maximum of four times (following~\cite{krizhevsky2012imagenet}). 
The networks were 
trained to 100,000--150,000 iterations, stopping when the 
validation error stopped improving even after 
decreasing the learning rate.

\subsection{Predicting Pairwise Comparisons}
\textbf{SS-CNN:} 
We experiment with SS-CNN initialized 
using AlexNet, PlacesNet, and VGGNet, and
evaluated their performance using three methods described below.

\begin{enumerate}
\item \textbf{Softmax:} We calculate the binary prediction accuracy of
  the softmax output for prediction of pairwise comparisons.
\item \textbf{TrueSkill:} We generate 30 ``synthetic'' pairwise
  comparisons per image using the network, by feeding random image
  pairs, and calculate the TrueSkill score for each image with these
  comparisons. We compare TrueSkill scores of the two images in a
  pair, to predict the winning
  image for each pair in the test set, and measure the binary
  prediction accuracy. We use this method since TrueSkill is able to
  generate stable scores for images, which allows us to reduce the
  noise in independent binary predictions on image pairs. 
\item \textbf{RankSVM:}  We feed a combined feature representation of
  the image pair obtained from the final convolution layer of SS-CNN 
  to a RankSVM~\cite{joachims2002optimizing} (using the 
  LIBLINEAR~\cite{fan2008} implementation), 
  and learn a ranking function. We then use the ranking scores 
  for images in the test set to decide the winner from test image 
  pairs, and calculate the binary prediction accuracy.
\end{enumerate} 

\begin{table}[t]
\begin{center}
\caption{Pairwise Comparison Prediction Accuracy}
\label{tab:res1}
\subfloat[SS-CNN]{%  
\begin{tabular}{|l|c|c|c|}
\hline
\multirow{2}{*} {\textbf{Network}}
&\multicolumn{3}{c|}{\textbf{Ranking Method}} \\ \cline{2-4}
&Softmax&TrueSkill&RankSVM\\ 
\hline
AlexNet & 53.0\% & 55.7\% & 58.4\% \\
SS-CNN (AlexNet) & 60.3\% & 62.6\% & 65.5\% \\
\hline
PlacesNet & 56.4\% & 58.8\% & 61.6\% \\
SS-CNN (PlacesNet) & 62.2\% & 64.7\% & 68.1\% \\
\hline
VGGNet  & 60.9\% & 62.7\% & 63.5\% \\
SS-CNN (VGGNet) & 65.3\% & 67.8\% & \textbf{72.4\%} \\ 
\hline
\end{tabular}
 }
\subfloat[RSS-CNN]{%
\renewcommand{\arraystretch}{2}
\begin{tabular}{|l|c|}
\hline
\textbf{Model} & \specialcell{\textbf{Prediction} \textbf{Acc.}} \\ \hline
%\textbf{Model} & \textbf{Prediction} \textbf{Acc.} \\ \hline 
AlexNet & 64.1\% \\ \hline
PlacesNet\hspace{0.2cm} & 68.8\% \\ \hline
VGGNet & \textbf{73.5\%} \\ \hline
\end{tabular}
\renewcommand{\arraystretch}{1}
 }
\end{center}
\vspace{-0.8cm}
\end{table}

\hspace{-0.55cm}We evaluate the accuracy for all three networks 
with (i) original (pre-trained) 
weights, and (ii) weights fine-tuned with the Place Pulse 2.0  dataset. 
Table~\ref{tab:res1}-(a) shows that, in all cases, the
binary prediction accuracy increases significantly---$6.5\%$ on
average---across all experiments. 
The gain in performance can be attributed to both, end-to-end
learning of the pairwise classification task 
and the size and diversity of the Place Pulse 2.0  dataset.
SS-CNN (VGGNet), the deepest architecture, obtains the best
performance over all three methods. 
We also observe that RankSVM
consistently outperforms TrueSkill, which in turn, outperforms 
softmax. This makes sense, since TrueSkill is not designed
to maximize prediction accuracy for pairwise comparisons, but rather to
generate stable ranked scores from pairwise comparisons. In contrast, 
the RankSVM loss function explicitly tries to minimize
misclassification in pairwise comparisons.
\if 0
It is notable that the pre-trained AlexNet features perform 
close to chance (58.5\% vs 50\%), even with a strong classifier like a
RankSVM, while the fine-tuned network, has an accuracy of  
We find that the \textit{fusion} network is successful in learning
discriminative features for pairwise comparisons, by identifying
context with the scene and object information. The 
context between scene and object information is absent in networks
trained just for object or scene classification. 
 The \textit{fusion} layers of SS-CNN allow us to learn this missing
contextual information.\fi
\\\\
\textbf{RSS-CNN:} We test the performance of the RSS-CNN
architecture with AlexNet, PlacesNet, and VGGNet. Since we explicitly
learn a ranking function $f_r(x)$ in the case of RSS-CNN, we
compare the ranking function outputs for both images in a test pair to
decide which image wins, and calculate the binary prediction accuracy. 
Table~\ref{tab:res1}-(b) summarizes the results for the three
models. The Ranking SS-CNN (VGGNet) obtains the highest accuracy for
pairwise comparison prediction (73.5\%). Since the RSS-CNN performs
end-to-end learning based on both the classification and ranking loss,
it significantly outperforms the SS-CNN trained with 
only classification loss (Table~\ref{tab:res1}-(a), column 1). The
RSS-CNN also does better than the combination of SS-CNN and 
RankSVM (Table~\ref{tab:res1}-(a), column 3) in most cases. 
We also find that RSS-CNN learns better with more data, and
continues to do so, whereas the SS-CNN architecture plateaus after
encountering approximately 60\% of the training data.

\begin{table}[t]
\begin{center}
\caption{Comparing Place Pulse 1.0 and Place Pulse 2.0 Datasets}
\vspace{0.2cm}
\label{tab:pp1}
\begin{tabular}{|l|c|c|c|}
\hline
Ranking SS-CNN & AlexNet & PlacesNet & VGGNet  \\ 
\hline
Place Pulse 1.0 (PP 1.0) & 59.8\% & 60.9\% & 64.1\% \\
Place Pulse 2.0  (same \#comparisons as PP 1.0) & 61.9\% & 66.2\% & 64.2\% \\
Place Pulse 2.0  (all comparisons) & 64.1\% & 68.8\% & 73.5\% \\
\hline
\end{tabular} 
\end{center}
\vspace{-0.4cm}
\end{table}

\subsection{Comparing Place Pulse 1.0 and Place Pulse 2.0 Datasets}
The Place Pulse 2.0 (PP 2.0) dataset has significantly higher 
visual diversity (56 cities from 28 countries) as compared to the
Place Pulse 1.0 (PP 1.0) dataset (4 cities from 2 countries). 
It also contains significantly more 
training data. For the visual attribute of 
\textit{Safety}, the PP 2.0  
dataset contains 370,134 comparisons for 110,988 images,
while the PP 1.0 dataset contains 73,806 comparisons
for 4,109 images. We are interested in studying the gain in 
performance obtained by this increased visual diversity and size.  
So we compare the binary prediction accuracy on PP 2.0 data, of an RSS-CNN 
trained with the three network architectures using
(i) all 73,806 comparisons from PP 1.0, (ii) 73,806 
comparisons randomly chosen from PP 2.0 (the same
amount of data as PP 1.0, but an increase in visual diversity), and 
(iii) 240,587 comparisons from PP 2.0 (the entire training set)   
(an increase in both the amount and the visual
diversity of data). Comparing
experiments (i) and (ii) (Table~\ref{tab:pp1}), we find  
that increasing visual diversity improves the accuracy
for all three networks, for the same amount of data. The gain in
performance is least for VGGNet, which is the deepest
network, and hence needs larger amount of data to train. 
Finally, training with the entire PP 2.0 dataset 
(experiment (iii)) improves accuracy by an average of
$7.2\%$ as compared to training with the PP 1.0. 

We also conduct the reverse experiment to measure the performance of
the PP 2.0 dataset on PP 1.0. We calculate the five-fold cross-validation
accuracy (following~\cite{porzi2015predicting}) for pairwise
comparison prediction for the \textit{Safety} attribute using 
a RankSVM trained with features of image pairs from 
the PP 1.0 dataset. We experiment with two different features,  
extracted, respectively, from (i) the SS-CNN (VGGNet) trained
with PP 2.0 data and (ii) the SS-CNN (VGGNet) trained with PP 2.0 data and
fine-tuned further with PP 1.0 data. Experiments (i) and (ii) yield an 
accuracy of $81.6\%$ and $81.1\%$ respectively. The
previous best result reported for the pairwise comparison prediction 
task~\cite{porzi2015predicting}
on the PP 1.0 dataset is $70.2\%$, albeit from a model trained with
PP 1.0 data alone. Note that our models are too deep to be trained with
only PP 1.0 data. 

\begin{table}[t]
  \begin{center}
\caption{Prediction Performance Across Attributes}
\label{tab:attributes}
\vspace{0.2cm}
    %\hspace{.5cm}%
        \begin{tabularx}{0.99\textwidth}{|l|Y|Y|Y|Y|Y|Y|}
          \hline
          %\noalign{\smallskip}
          \backslashbox{Train}{Test}  & Safe & Lively & Beautiful &
          Wealthy & Boring & Depressing \\ 
          \hline
          Safe & \textbf{73.5\%} & 67.7\% & 66.3\% & 60.3\% & 47.2\% & 42.3\%\\
          Lively & 63.8\% & \textbf{70.3\%} & 65.8\% & 61.3\% & 58.9\% & 53.7\% \\
          Beautiful & 61.2\% & 67.1\% & \textbf{70.2\%} & 53.5\% & 50.2\% & 51.4\% \\
          Wealthy & 60.7\% & 54.6\% & 52.7\% & \textbf{65.7\%} & 52.8\% & 55.9\%\\
          Boring & 48.6\% & 55.6\% & 52.3\% & {53.1\%} & \textbf{66.1\%} & 59.8\%\\
          Depressing & 54.5\% & 54.2\% & 43.2\% & 49.7\% & 57.2\% & \textbf{62.8\%}\\

          \hline
%          \textbf{All} & 73.1\% & {70.1\%} & 69.2\% &
%          \textbf{67.6\%}\\ 
%          \hline
        \end{tabularx}
 \end{center}
\vspace{-0.4cm}
\end{table}

\subsubsection{Comparison With Generic Image Features:} Prior
work~\cite{naik2014streetscore,ordonez2014learning,porzi2015predicting}
has found that generic image features do well on the Place Pulse
1.0 dataset, for predicting both ranked scores and pairwise 
comparisons. Based on this literature, we extract three best
performing features---GIST~\cite{oliva2001modeling},
Texton Histograms~\cite{malik2001contour}, and CIELab Color 
Histograms~\cite{xiao2010sun}---from images in the PP 2.0 dataset. 
We find that the pairwise prediction 
accuracy of a RankSVM trained with feature vector consisting 
of these features is $56.7\%$ on the PP 2.0 dataset, 
significantly lower than all
variations of SS-CNN. Our  best performing model 
RSS-CNN (VGGNet) has an accuracy of $73.5\%$.

\subsection{Predicting Different Perceptual Attributes}
Our dataset contains a total of six perceptual attributes---\textit{Safe}, \textit{Lively}, \textit{Beautiful},
\textit{Wealthy}, \textit{Boring}, and \textit{Depressing}. We now
evaluate the prediction performance of RSS-CNN on these six
attributes. Specifically, we train the 
RSS-CNN (VGGNet) network for each attribute, and
measure it's performance using binary prediction accuracy. 
Table~\ref{tab:attributes} shows that the
\textit{in-attribute} prediction performance is roughly 
proportional to the
number of comparisons available for training, with the best prediction
performance for \textit{Safe}, and the worst performance for
\textit{Depressing}. 
We also evaluate the performance of the network trained to predict
one perceptual attribute in predicting the pairwise comparisons for
the other three attributes (\textit{cross-attribute} prediction). 
The \textit{Safe} network shows strong
performance in prediction of \textit{Lively}, \textit{Beautiful}, and
\textit{Wealthy} attributes, which is indicative of the high
correlation between different perceptual attributes. 

\begin{figure*}[t]
    \centering
    \includegraphics[width=0.9\textwidth]{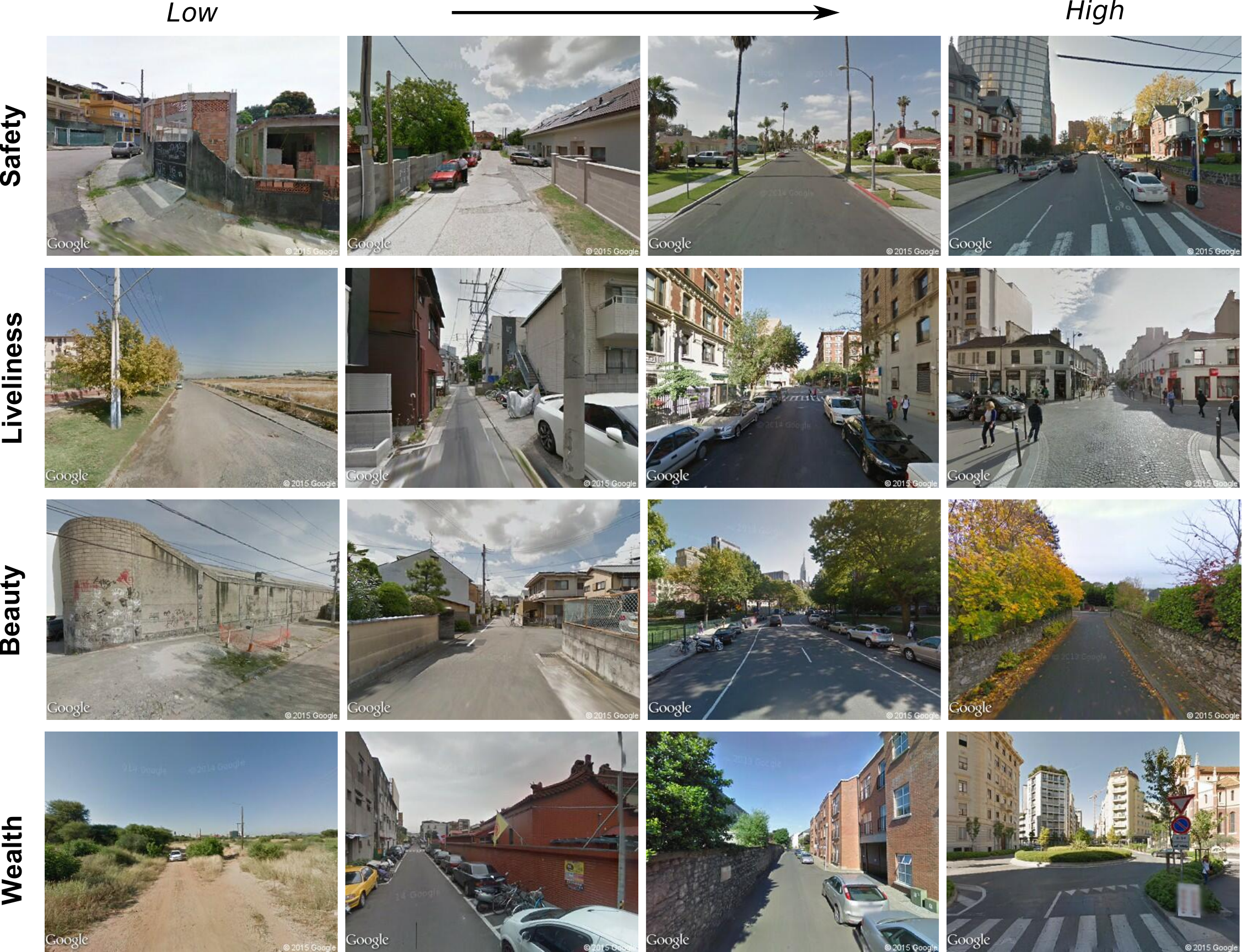} 
   \caption{Example results from the Place Pulse 2.0 dataset,
     containing images ranked based on pairwise comparisons generated 
     by the RSS-CNN.}
\vspace{-0.4cm}
\label{fig:data_samples}
\end{figure*}

\begin{figure*}[t]
    \centering
    \includegraphics[width=0.95\textwidth]{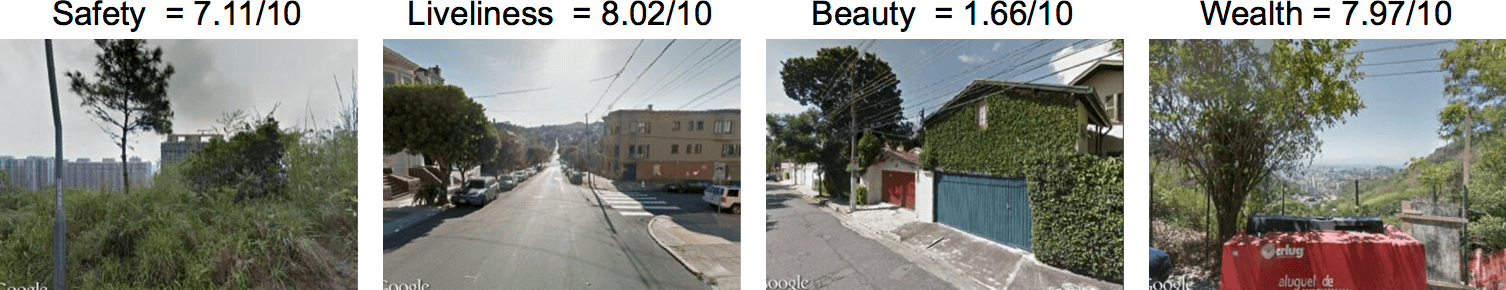} 
   \caption{Example failure cases from the prediction results,
     containing images and their TrueSkill scores for attributes
     computed from pairwise comparisons generated by the RSS-CNN.} 
\vspace{-0.4cm}
\label{fig:failure}
\end{figure*}

\if 0
Next, we evaluate the performance
of a network trained using pairwise comparisons for \textit{all} four
attributes, which amount to a total of 943,139
comparisons. The
additional data does not improve the accuracy in case of the
\textit{Safe},\textit{lively}, and \textit{Beautiful} attributes, 
indicating that these
attributes do not gain much from the pairwise comparisons from the
other attributes. However, this network improves prediction 
performance for \textit{Wealthy} by $1.9\%$.     
\fi
A model trained to predict pairwise comparisons can be used to
generate ``synthetic'' comparisons by taking random image pairs as
input. A large number of 
comparisons can be then fed to ranking algorithms (like
TrueSkill) to obtain stable ranked scores. We use this trick 
to generate TrueSkill scores for four attributes
using pairwise comparisons predicted by a trained RSS-CNN (VGGNet) 
(30 per image). Figure~\ref{fig:data_samples} shows examples from the
dataset, and  figure~\ref{fig:failure} shows failure cases. 
We find that, for instance, highway images with forest cover
are predicted to be highly safe, and overcast images as highly boring.  
Quantitatively, the correlation coefficient ($R^2$) of
\textit{Safe} with \textit{Lively}, \textit{Beautiful}, and 
\textit{Wealthy} is $0.80$, $0.83$, and $0.65$ respectively. This
indicates that there is relatively large orthogonality ($(1-R^2)$) 
between attributes.

%Places perceived as safe are also more likely to be
%perceived as wealthy 
%(Fig. 3D R2= 68.94%, p-value,0.0001)
%and unique (Fig. 3E R2= 35.32%, p-value,0.0001), yet, their
%orthogonal components (1-R2
%) are relatively large

\subsection{Predicting Urban Appearance across Countries}
Our hope is that the Place Pulse 2.0 dataset will enable 
algorithms to conduct automated audits of 
urban appearance for cities all
across the world. The Streetscore~\cite{naik2014streetscore} 
algorithm was able to successfully generalize to the Northeast and
Midwest of the U.S.,
based on training data from just two cities, New York and Boston.
This indicates that models trained with the PP 2.0 dataset 
containing images from 28 countries should be able
to generalize to large regions in these countries, and beyond. 
For a qualitative experiment to test generalization, 
we download 22,282 Street
View images from six cities from six continents---Vancouver, 
Buenos Aires,
St. Petersburg, Durban, Seoul, and Brisbane---that were not a 
part of the PP 2.0 dataset. We
map the perceived safety for these cities using TrueSkill scores 
for images computed from 30 ``synthetic'' 
pairwise comparisons generated with RSS-CNN (VGGNet). While the
prediction performance of the network on these images cannot be
quantified due to a lack of human-labeled ground truth, 
visual inspection shows that the scores assigned to
streetscapes conform with visual intuition (see supplement for
map visualizations and example images). 
%The average perceptual scores of these cities agree with the
%aggregate rankings of these continents (Figure~\ref{fig:global}, also
%see Supplement for comprehensive rankings).

\if 0
\begin{figure*}[t]
    \centering
    \includegraphics[width=0.95\textwidth]{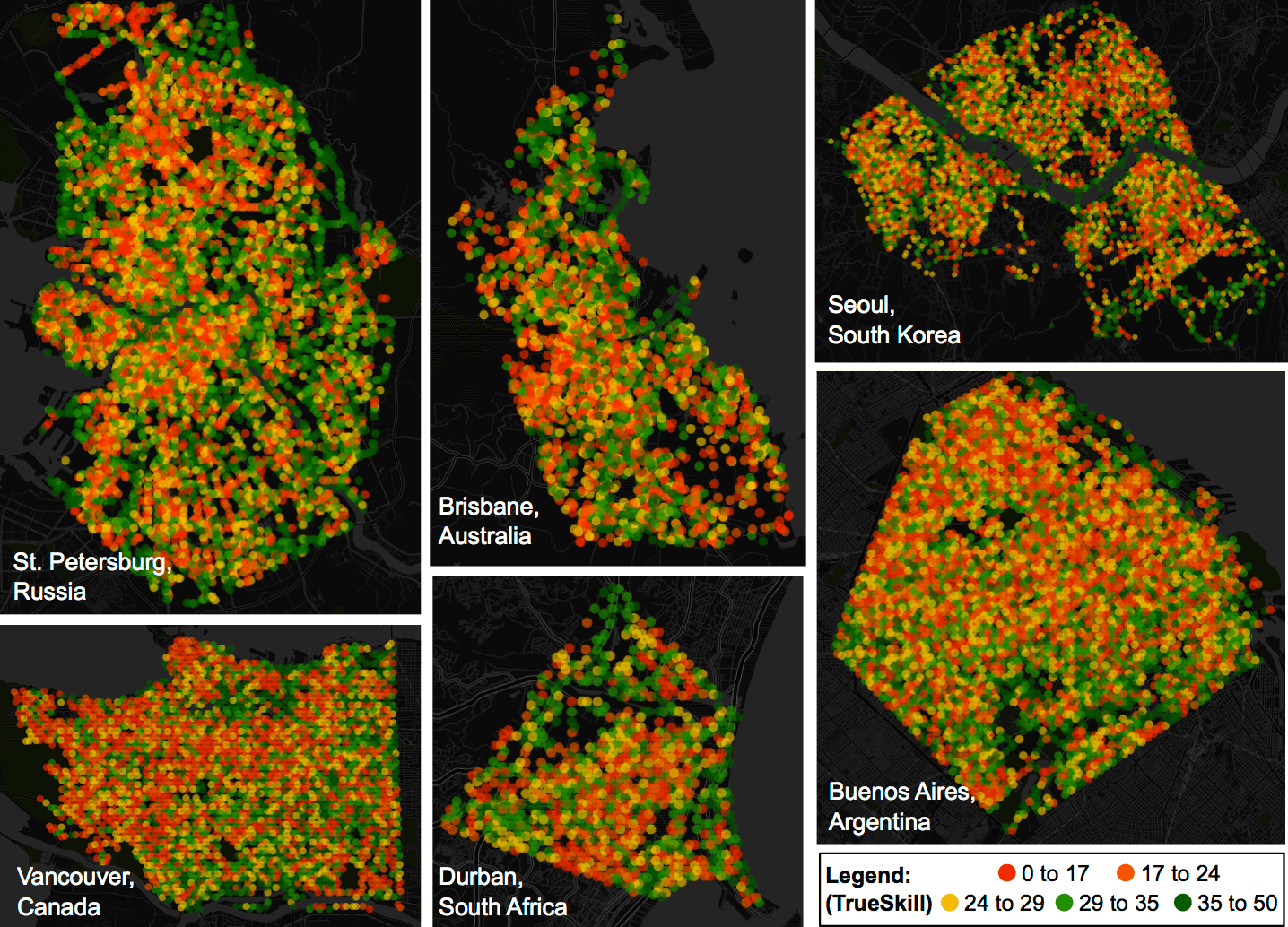} 
   \caption{We map Trueskill scores for \textit{safety} 
     for 6 cities (from 6
     continents) that were not a part of the Place Pulse 2.0 
     dataset, using pairwise comparisons generated by a 
     trained RSS-CNN. (Note: maps at different scales)} 
\vspace{-0.4cm}
\label{fig:global}
\end{figure*}
\fi

\section{Discussion \& Concluding Remarks}
In this paper, we introduced a new crowdsourced 
dataset of global urban appearance containing pairwise image comparisons
and proposed a neural network architecture for predicting the 
human-labeled comparisons. 
Since we focussed on predicting pairwise win/loss decisions to aid image
ranking, we ignored the image pairs where the users perceive the images
to be \textit{equal} for the given perceptual attribute. However,
$13.2\%$ pairwise comparisons in our dataset are \textit{equal}, and
incorporating the prediction of equality in comparisons should be a
part of future work. Future work 
can also explore the determinants of perceptual attributes of urban
appearance (e.g., what makes an image appear safe? or lively?)
Such studies would allow
better visual designs that optimize attributes of urban
appearance. From a computer vision perspective, understanding the
geographical range over which models trained on street-level imagery 
from different regions of the world are able to generalize would be an
interesting future direction, since the architectural similarities
between cities are determined by a complex interaction of history,
culture, and economics. 

Our technique can be generalized for computer vision tasks
of studying the style, perception, or visual attributes of images,
objects, or scene categories. Our trained networks can be used to 
generate a global dataset of urban appearance, which will enable
the study a variety of research questions: 
How does urban appearance affect the behavior and health of
residents, and how do these effects vary across countries? 
How are different
architectural styles perceived? How similar/different are
different cities across the world in terms of perception? Can visual
appearance be used as a proxy for inequality within cities?
A global dataset of urban
appearance will thus aid computational studies in 
architecture, art history, sociology, and economics.         
These datasets can also help policymakers and 
city governments make data-driven decisions on allocation
of resources to different cities or neighborhoods 
for improving urban appearance. 

\subsection*{Acknowledgements}

We gratefully thank Abhishek Das, Arjun Chandrasekaran and Deepak 
Jagdish for the inputs and assistance at various stages in this work.

%\clearpage

\bibliographystyle{splncs}
\bibliography{refs,social}

\title{\hspace*{-1pt}Supplementary Material: Deep Learning the City: Quantifying Urban Perception At A
  Global Scale} 
\titlerunning{Deep Learning the City : Supplementary Material}

\authorrunning{A. Dubey, N. Naik, D.Parikh, R. Raskar, and  C. Hidalgo}

\author{Abhimanyu Dubey\textsuperscript 1, Nikhil Naik\textsuperscript
  3, Devi Parikh\textsuperscript 2 \\ Ramesh Raskar\textsuperscript 3,
 C\'esar A. Hidalgo\textsuperscript 3}
 \institute{\textsuperscript 1 Indian Institute of Technology Delhi\\
  \email{abhimanyu1401@gmail.com}\\
  \textsuperscript 2 Virginia Tech\\
  \email{ parikh@vt.edu}\\
  \textsuperscript 3 MIT Media Lab\\
  \email{ \{naik,raskar,hidalgo\}@mit.edu}\\
}
\maketitle
\section*{Supplemental Contents}
\noindent This supplement is organized as follows:
\begin{itemize}[itemsep=1pt]
\item Section 1 contains analysis on training data size versus
  prediction performance. 
\item Section 2 contains discussion on the correlation between
  perceptual attributes.
\item Section 3 shows example images and their perceptual attributes
  for the Place Pulse 2.0 dataset, along with example images and 
maps from the six additional cities that were not a part of 
the training dataset.  
\end{itemize}

\section{Size of Training Data and Accuracy}
\begin{figure*}[h]
\centering
\begin{tabular}{ccc}
\includegraphics[width=0.3\textwidth]{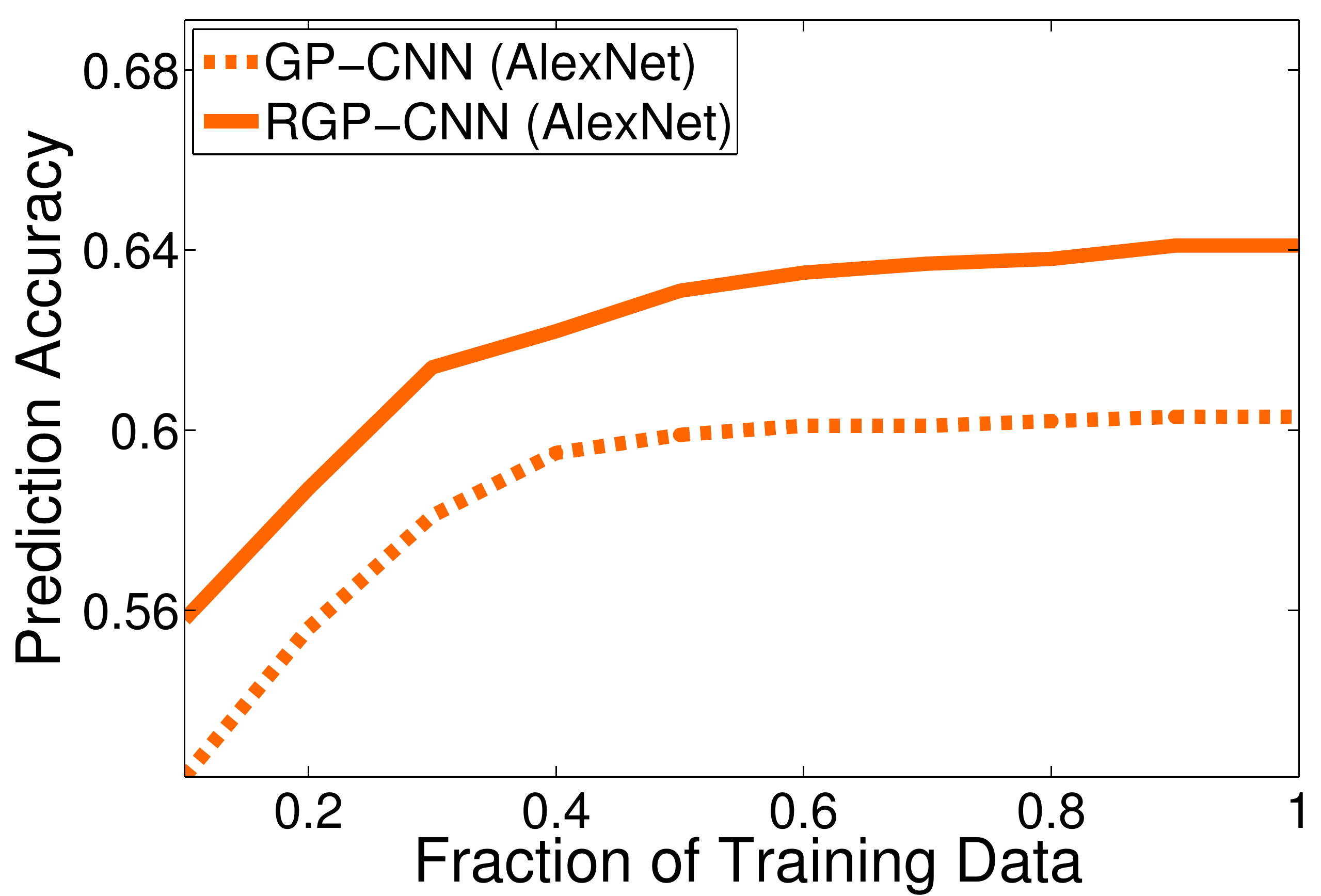}\hspace{0cm}&
\includegraphics[width=0.3\textwidth]{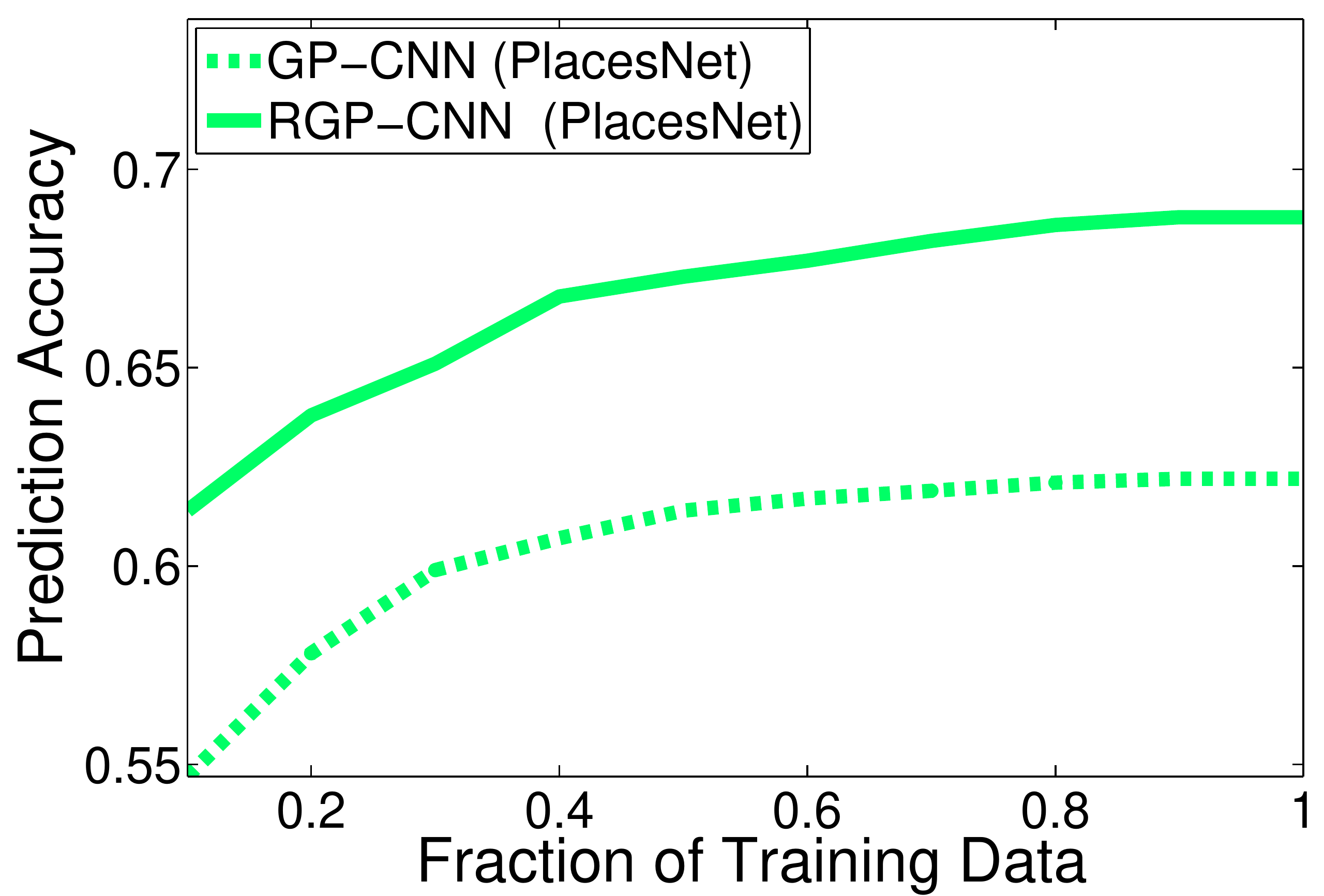}\hspace{0cm} &
\includegraphics[width=0.3\textwidth]{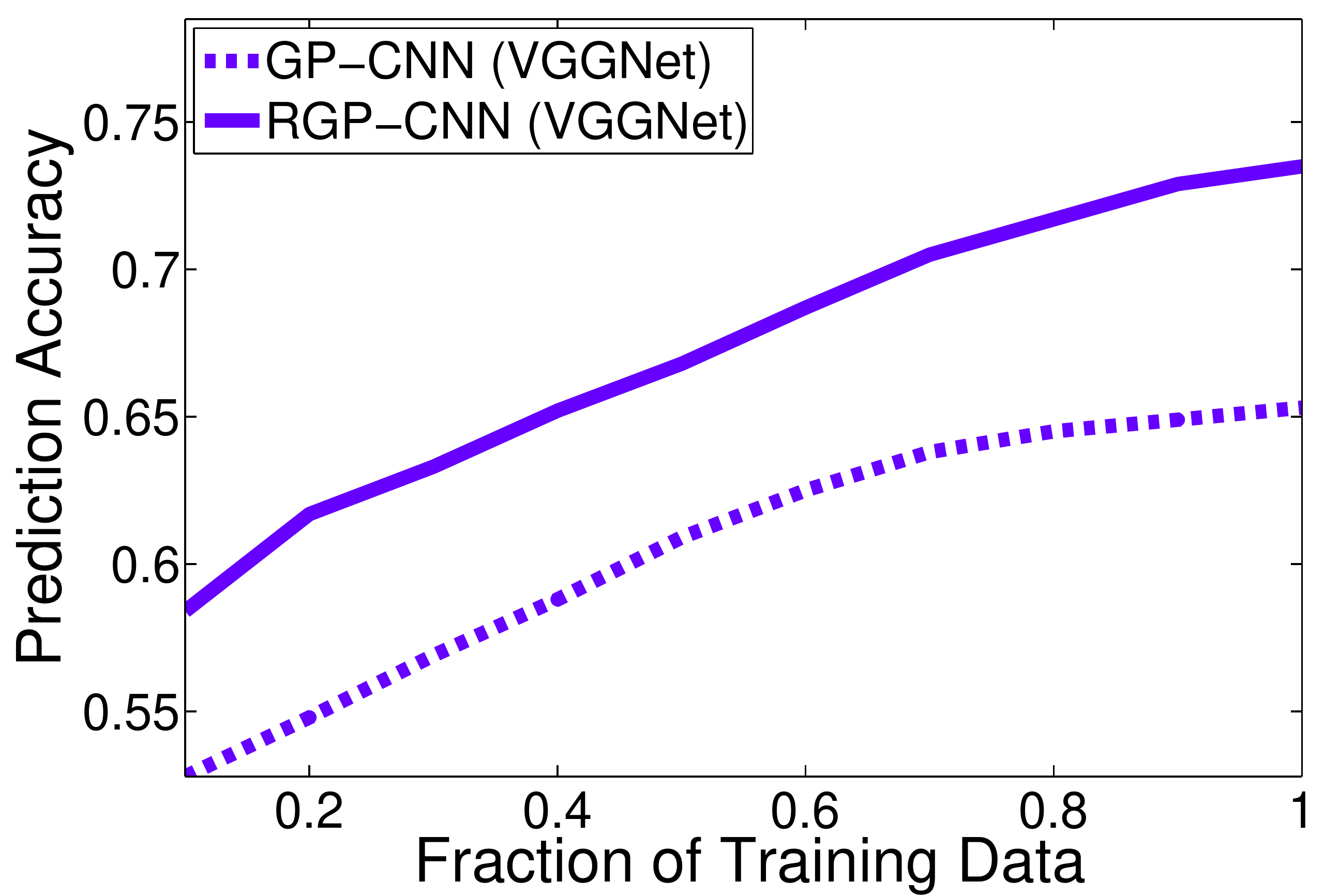}\hspace{0cm}\\
(a) & (b) & (c)
\end{tabular}
\caption{We plot the fraction of training data versus accuracy for
  SS-CNN and RSS-CNN evaluated with the three models.}
\label{fig:data_acc}
\end{figure*}

\noindent We evaluate the performance of SS-CNN and RSS-CNN for
different sizes of training data from the Place Pulse 2.0 (PP 2.0)
dataset, for the perceptual attribute of \textit{Safety}, which 
contains $240,587$ 
comparisons in the training set and $111,040$ comparisons in
the test set (and the rest in the validation set). We train SS-CNN and
RSS-CNN on fractions of training data, starting with $10\%$ of data,
and increasing the size in steps of $10\%$, and measure the
performance in the form of binary prediction accuracy, fine-tuning
from the three basic networks (AlexNet, PlacesNet, VGGNet). The
results (Figure~\ref{fig:data_acc}) show that in case of SS-CNN, the 
the performance plateaus after approximately $50\%$ of data for 
both AlexNet and PlacesNet, while there is a significant increase 
in accuracy for
VGGNet, likely due to the fact that the deeper network learns better
with more data. In case of RSS-CNN, the performance plateaus after
approximately $80\%$ of data for both AlexNet and PlacesNet, while there
is a quite steady increase in accuracy for VGGNet until $100\%$ of
training data is used. The difference in trends between SS-CNN and
RSS-CNN can be attributed to the additional learning capacity of 
ranking layers.  

\section{Correlation between Perceptual Attributes}
As discussed in Section 5.3 of the main text, we are interested in
understanding the orthogonality between the six perceptual
attributes (\textit{Safe}, \textit{Lively}, \textit{Beautiful},
\textit{Wealthy}, \textit{Boring}, and \textit{Depressing}). 
We generate TrueSkill scores for all images in the
PP 2.0 dataset, and measure the Squared 
Pearson Correlation Coefficient ($R^2$) between pairs of attributes
(Table~\ref{tab:attributes}). We find that the attribute \textit{Safe}
has the largest positive correlation with \textit{Beautiful}, and the
largest negative correlation 
with \textit{Boring}. The table 
demonstrates that the different perceptual attributes are
measuring qualities that are not highly correlated or redundant.     

\begin{table}[t]
  \begin{center}
\caption{Correlation between Perceptual Attributes}
\label{tab:attributes}
\vspace{0.2cm}
    %\hspace{.5cm}%
        \begin{tabularx}{.95\textwidth}{|l|Y|Y|Y|Y|Y|Y|}
          \hline
          %\noalign{\smallskip}
          $R^2$  & Safe & Lively & Beautiful & Wealthy & Boring & Depressing \\ 
          \hline
          Safe & 1.00 & 0.80 & 0.83 & 0.65 & -0.36 & -0.22 \\
          Lively & 0.80 & 1.00 & 0.71 & 0.68 & -0.71 & -0.42 \\
          Beautiful & 0.83 & 0.71 & 1.00 & 0.75 & 0.15 & -0.28\\
          Wealthy & 0.65 & 0.68 & 0.75 & 1.00 & 0.27 & -0.34\\
          Boring & -0.36 & -0.71 & 0.15 & 0.27 & 1.00 & 0.39 \\
          Depressing & -0.22 & -0.42 & -0.28 & -0.34 & 0.39 & 1.00 \\
          \hline
        \end{tabularx}
 \end{center}
\vspace{-0.4cm}
\end{table}

\section{Example Images and Perceptual Attributes}
\begin{figure*}[t]
\centering
\includegraphics[width=1.0\textwidth]{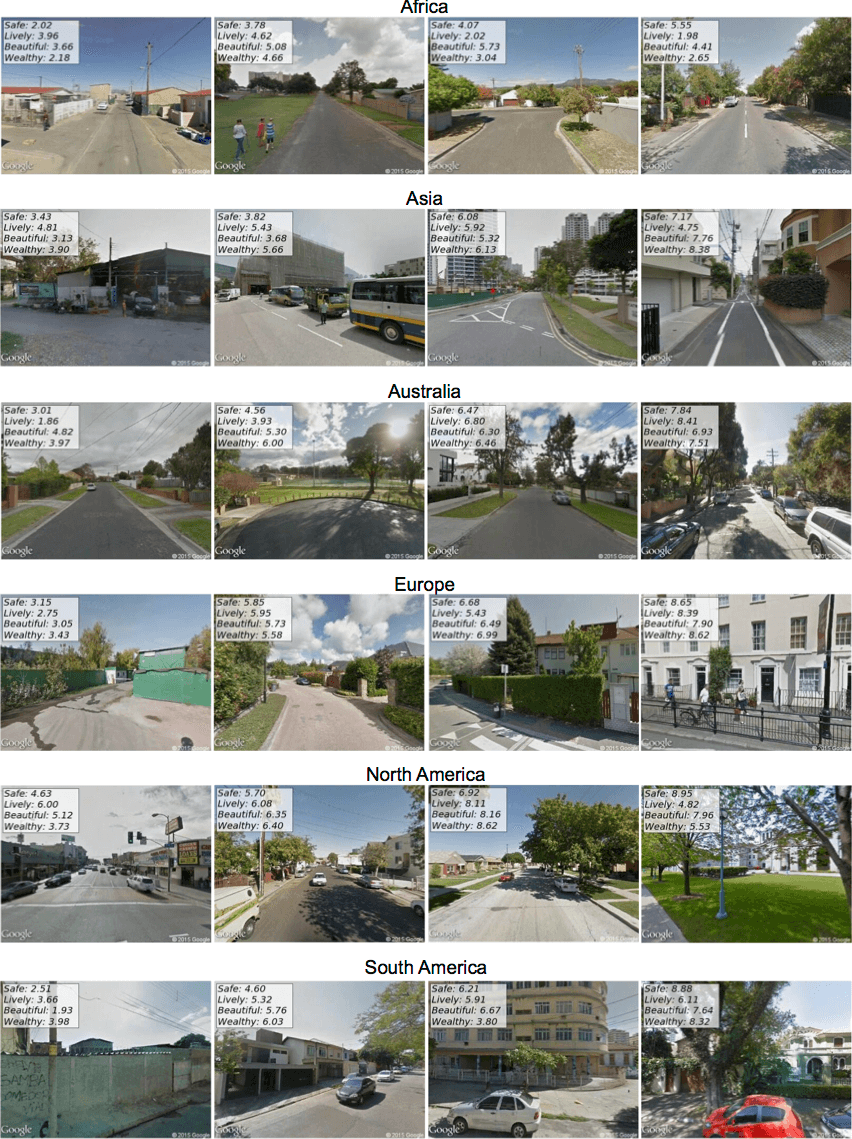}
\caption{Image examples, with their perceptual attributes, from all
  six continents from the Place Pulse 2.0 dataset (all scores out of
10).}
\label{fig:all}
\end{figure*}

The Place Pulse 2.0 (PP 2.0) dataset contains significant 
visual diversity, with images from 56 cities from 28 countries 
spread across 6 continents. After training RSS-CNN (VGGNet), 
we generate 30 ``synthetic'' pairwise
comparisons for each image in the SS dataset, 
by feeding randomly selected image pairs
to this network. We use these comparisons to generate
TrueSkill scores~\cite{herbrich2006trueskill} for all images. 
Each images' TrueSkill is 
modeled as a $\mathcal{N}(\mu,\sigma^2)$\ random variable, which gets
updated after every contest.
The TrueSkills for players $x$ and $y$ in a two-player 
contest in which $x$ wins against $y$, are updated as, 
\vspace{-0.05in}
\begin{equation}
\begin{gathered}
\mu_x\longleftarrow\mu_x +
\frac{\sigma^2_x}{c}\cdot f\left(\frac{\left(\mu_x-\mu_y\right)}{c},\frac{\varepsilon}{c}\right),
\\
\mu_y\longleftarrow\mu_y -
\frac{\sigma^2_y}{c}\cdot f\left(\frac{\left(\mu_x-\mu_y\right)}{c},\frac{\varepsilon}{c}\right), 
\\
\sigma^2_x\longleftarrow\sigma^2_x\cdot\left[1 - 
\frac{\sigma^2_x}{c}\cdot g\left(\frac{\left(\mu_x-\mu_y\right)}{c},\frac{\varepsilon}{c}\right)
\right],
\\
\sigma^2_y\longleftarrow\sigma^2_y\cdot\left[1 - 
\frac{\sigma^2_y}{c}\cdot g\left(\frac{\left(\mu_x-\mu_y\right)}{c},\frac{\varepsilon}{c}\right)
\right],
\\
c^2 = 2\beta^2 + \sigma_x^2 + \sigma_y^2,
\label{eqn:ts_update}
\end{gathered}
\end{equation}

\noindent where $\mathcal{N}(\mu,\sigma_x^2)$\ and $\mathcal{N}(\mu,\sigma_y^2)$\ are
TrueSkills of $x$ and $y$. The constant $\beta$ 
represents a per-game variance, and $\varepsilon$ is the empirically 
estimated probability that two players will tie. Functions  
$f\left(\theta\right) = \mathcal{N}(\theta)/\Phi(\theta)$ and
$g(\theta) = f(\theta)\cdot(f(\theta) + \theta)$ are defined
using the Normal PDF $\mathcal{N}(\theta)$ and Normal 
CDF $\Phi(\theta)$. Following ~\cite{herbrich2006trueskill}, 
we use ($\mu=25, \sigma=25/3$) as initial values for
rankings for all images and choose $\beta =
25/3$ and $\varepsilon = 0.1333$. After all updates are completed, we
use only the $\mu$ as the TrueSkill score for each image, and scale
the scores to a range between $0$ and $10$. Note that the TrueSkill score
generation process is similar to previous
work (e.g.,~\cite{naik2014streetscore,kiapour2014hipster})---it is
reproduced here for clarity. Figure~\ref{fig:all} shows example images
and their TrueSkill scores for all four attributes, generated with the
process described above. 

\begin{figure*}[t]
\centering
\includegraphics[width=1.0\textwidth]{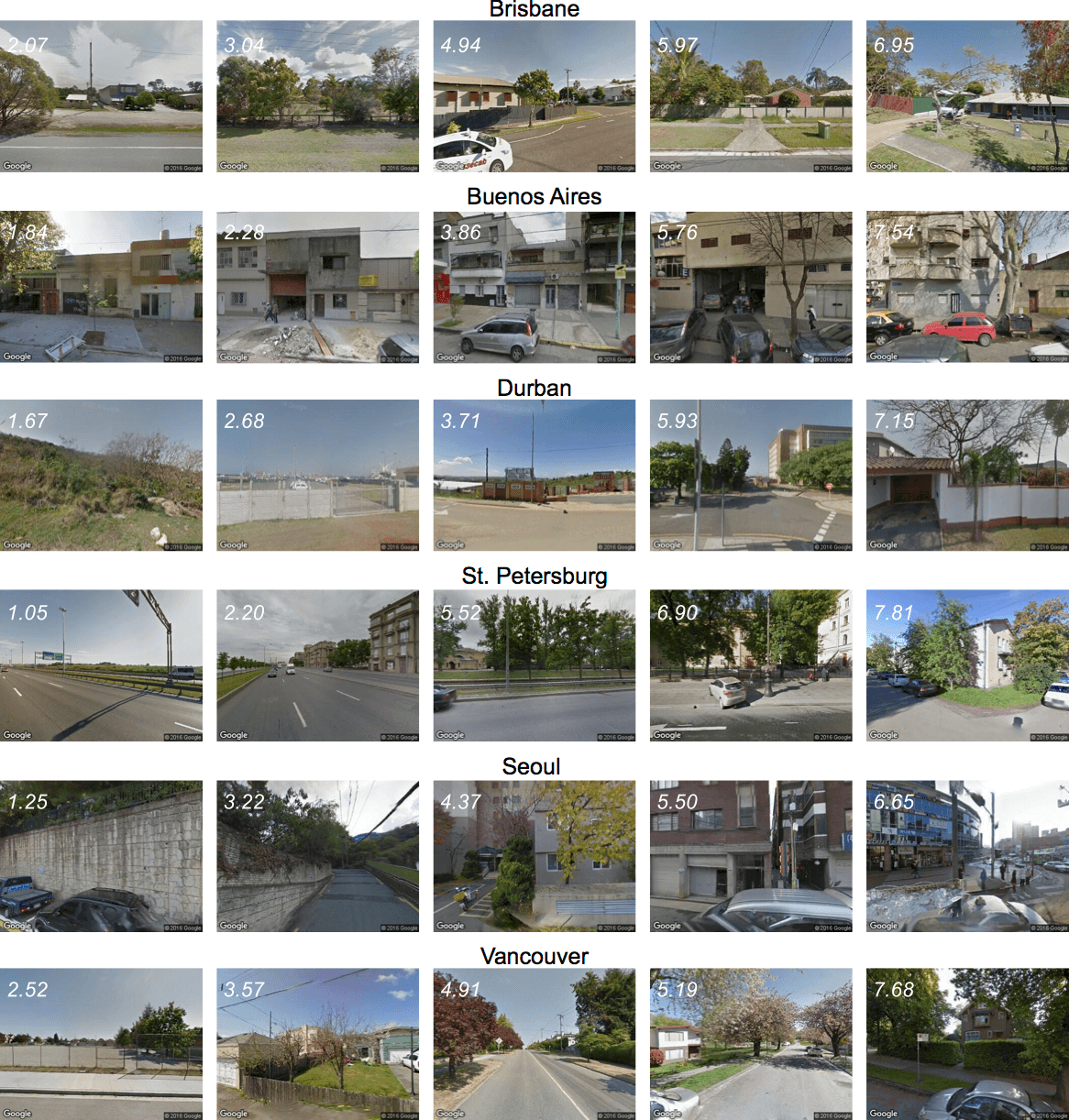}
\caption{Image examples, with their score for perceived safety, 
from six cities that were not a part of the Place Pulse 2.0 dataset 
(all scores out of 10).}
\label{fig:extrap}
\end{figure*}

Similarly, we generate TrueSkill scores for perceived safety 
for images from six cities that were not a
part of the PP 2.0 dataset (Section 5.4 of the main text). We generate 
30 pairwise comparisons for each image. 15 of the 30 comparisons are
generated from image pairs where first image is from the PP 2.0 dataset,
and the second image is from one the new cities. The remaining 15
comparisons are generated from image pairs where both images are from
the new cities (chosen randomly). Figure~\ref{fig:extrap} shows
example images and their TrueSkill scores, which conform with visual
inspection. 

\begin{figure*}[t]
    \centering
    \includegraphics[width=0.95\textwidth]{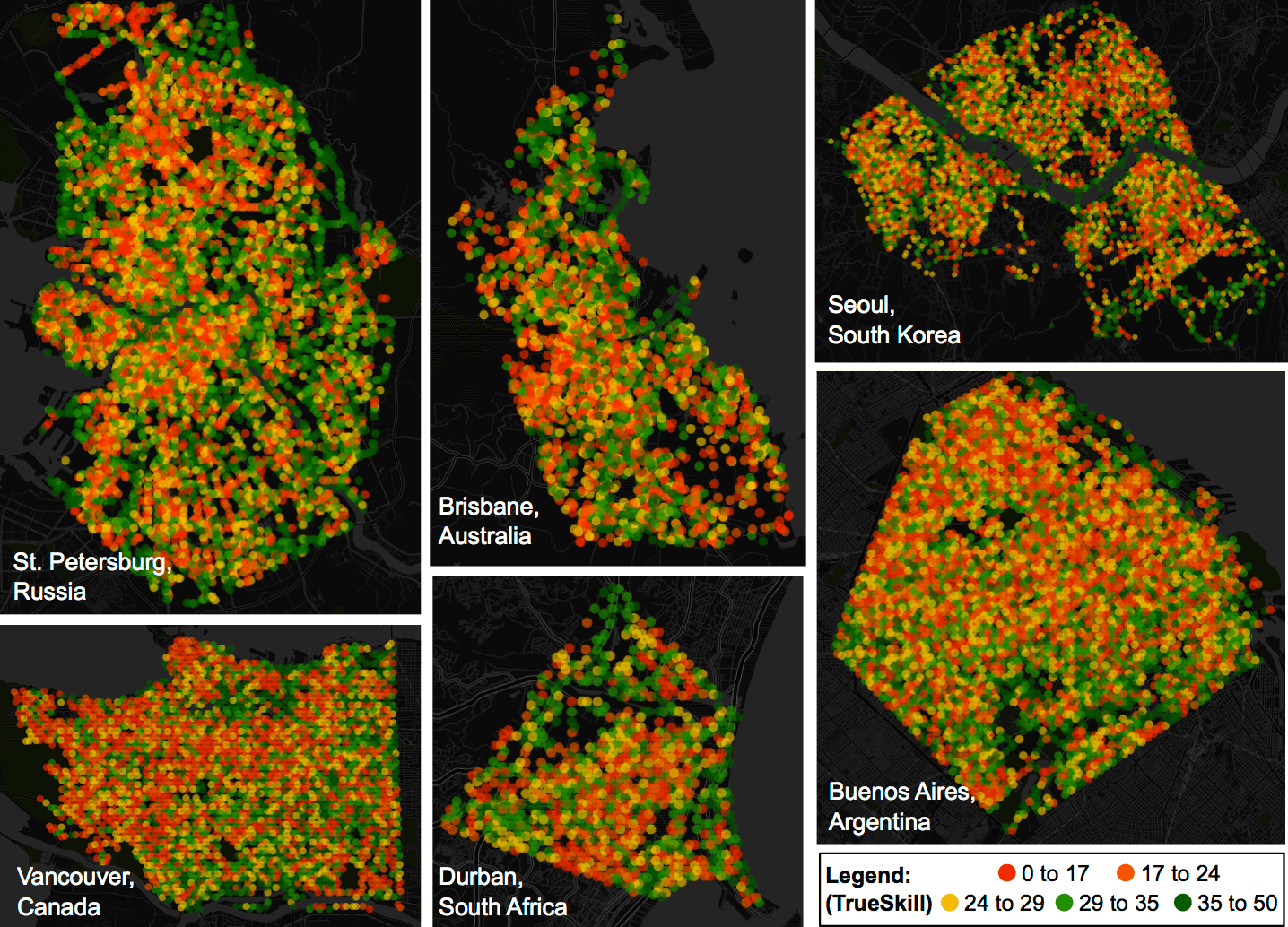} 
   \caption{We map TrueSkill scores for \textit{safety} 
     for 6 cities (from 6
     continents) that were not a part of the Place Pulse 2.0 
     dataset, using pairwise comparisons generated by a 
     trained RSS-CNN. (Note: maps at different scales)} 
\vspace{-0.4cm}
\label{fig:global}
\end{figure*}

\end{document}